\documentclass[10pt,twocolumn,letterpaper]{article}

\usepackage{iccv}
\usepackage{times}
\usepackage{epsfig}
\usepackage{graphicx}
\usepackage{amsmath}
\usepackage{amssymb}
\usepackage{graphicx}
\usepackage{algorithm}
\usepackage{algorithmic}
\usepackage{makecell}
\usepackage{multicol}
\usepackage{color}
\usepackage{multirow}
\usepackage{xcolor}

\usepackage[breaklinks=true,bookmarks=false]{hyperref}

\iccvfinalcopy 

\def\iccvPaperID{42} 

\ificcvfinal\fi

\begin{document}

\title{RCD-SGD: Resource-Constrained Distributed SGD in Heterogeneous Environment Via Submodular Partitioning}

\author{Haoze He\thanks{Corresponding author.}\\
New York University, NY, USA\\
{\tt\small hh2537@nyu.edu}
\and
Parijat Dube\\
IBM Research, NY, USA\\
{\tt\small pdube@us.ibm.com}
}

\maketitle
\ificcvfinal\thispagestyle{empty}\fi

\begin{abstract}
The convergence of SGD based distributed training algorithms is tied to the data distribution across workers. Standard partitioning techniques try to achieve equal-sized partitions with per-class population distribution in proportion to the total dataset. Partitions having the same overall population size or even the same number of samples per class may still have Non-IID distribution in the feature space. In heterogeneous computing environments, when devices have different computing capabilities, even-sized partitions across devices can lead to the straggler problem in distributed SGD. We develop a framework for distributed SGD in heterogeneous environments based on a novel data partitioning algorithm involving submodular optimization. Our data partitioning algorithm explicitly accounts for resource heterogeneity across workers while achieving similar class-level feature distribution and maintaining class balance. Based on this algorithm, we develop a distributed SGD framework that can accelerate existing SOTA distributed training algorithms by up to 32\%.  
\end{abstract}

\section{Introduction}
\label{sec:intro}
Stochastic gradient descent (SGD) is the skeleton of most state-of-the-art (SOTA) machine learning algorithms. Traditional SGD was designed to be run serially at a single worker. However, with the increasing size of deep learning models and dataset, too much time is required to perform training using a single machine~\cite{devlin2018bert}. Parallelism in training is inevitably necessary to deal with this magnitude of model, data, and compute requirements~\cite{narayanan2021efficient}.
Distributed SGD parallelizes the training across multiple workers, through different types of parallelism (model~\cite{wang2022fine}, data~\cite{wang2019matcha,lian2017can}, pipeline) to speed up training. While parallelism can achieve high training throughput, guaranteeing the stability and convergence of SGD is challenging in distributed training~\cite{lian2018asynchronous, wang2019matcha}. It is a complex interplay of machine learning hyperparameters and dataset distribution that contributes to convergence of distributed machine learning training. Most of the SOTA decentralized distributed machine learning frameworks~\cite{lian2017can, wang2019matcha,blot2016gossip, jin2016scale, lian2018asynchronous} randomly partition the original dataset into subsets and assign them to different workers. Their convergence analysis simply assumes the subsets in each worker are independent and identically distributed (IID). However, random partitioning cannot guarantee IID at the feature level and the Non-IID issue still exists.

Submodular optimization, associated with a rich family of submodular functions that can measure the diversity of a given subset, has been used in machine learning in the past decade. Submodular optimization is well known for its strong capability to select a representative subset. Earlier work have tried applying submodular optimization to real-world applications including speech recognition, active learning, and computer vision \cite{jegelka2011submodularity,krause2008near,nagano2010minimum,krause2008robust,liu2013submodular,wei2013using,wei2014submodular,wei2015submodularity,zheng2014submodular}. To solve the Non-IID issue in distributed machine learning, \cite{wei2015intelligently} tries to use submodular optimization to partition the original dataset into IID subsets. General partition on any dataset ahead of training won’t give additional computational costs and will lead to faster convergence. However, the submodular partition algorithm GreedMax \cite{wei2015intelligently} will lead to different sized subsets after partition. Different sized subsets will lead to straggler problem (the delays in waiting for the learners with large batch size) due to synchronization step in most distributed SGD. Straggler problem" refers to delays caused by the slowest learners in distributed machine learning. This problem is caused by the ”synchronization barrier,” which is the process of coordinating updates to the model parameters across multiple machines or devices, typically done after each iteration of the training process~\cite{tandon2017gradient,lian2018asynchronous}. Although some recent papers try to explore the possibility of using submodular optimization to partition dataset under certain constraints~\cite{wang2021constrained}, it cannot be applied to distributed machine learning directly due to non-uniformly sized and imbalanced classes across partitions. To solve these problems, we propose RCD-SGD, a resource-constrained distributed SGD. Our main contributions are: 
\begin{enumerate}
    \item  We propose a novel algorithm to partition data for distributed SGD in heterogeneous environments. The proposed algorithm partitions the dataset across workers in proportion to their computational capabilities while achieving similarity in class-level feature distribution and maintaining class balance. (Class-level feature distribution refers to the pattern of specific feature values within different categories or classes in a dataset, providing insights into the data's characteristics and feature importance. Maintaining class data sample number balance means ensuring that the number of data samples in each class or category of a dataset is equal.) Our algorithm reduces the computational complexity of earlier partitioning algorithms by a factor proportional to the number of classes in the dataset \cite{wang2021constrained}.
    \item By partitioning the original dataset into subsets with the same number of data points, the label-balanced greedy partition algorithm addresses the straggler problem in distributed synchronous SGD. 
    \item We evaluate the RCD-SGD algorithm using two different submodular functions and two SOTA-distributed SGD algorithms. By achieving IID partitioning of the dataset our algorithm achieves faster convergence than the SOTA baseline. With approximately local IID subsets, we reduce the communication frequency of distributed SGD and achieve up to 32\% speedup in wall-clock time when compared with the SOTA algorithms~\cite{stich2018local}. The final model also achieves slightly better loss and improves the final accuracy by 1.1\% ("Approximately local IID subsets" means subsets of data partitioned from original dataset are expected to be similar and follow similar distribution properties). 
    \item The proposed resource-constrained distributed SGD is a general algorithm that is extendable to most distributed SGD SOTA algorithms. Using any algorithm as a baseline, resource-constrained distributed SGD can achieve faster convergence and shorter wall-clock training time. 
\end{enumerate}


\section{Preliminaries and Related Work}

\subsection{Problem Formulation}

Distributed Stochastic Gradient Descent (SGD) aims to accelerate the training process by parallelizing it across multiple worker nodes, employing various types of parallelism such as model, data, and pipeline parallelism. Let us consider a network with $N$ worker nodes implementing distributed SGD. The model parameters are denoted by $x$, where $x \in \mathbb{R}^d$. Each worker node $i$ has access only to its own local training data, distributed as $D_i$. The objective of distributed SGD is to train a model by minimizing the objective function $L(x)$ using $N$ worker nodes. The challenge of distributed SGD can be formulated as follows:

\begin{equation}
\min_{x \in \mathbb{R}^d} L(x) = \min_{x \in \mathbb{R}^d} \frac{1}{N} \sum_{i=1}^{N} E_{s \sim D_i}[l_i(x;s)],
\end{equation}

where $l(x)$ is the loss function defined by the learning model, and $E_{s \sim D_i}[l_i(x;s)]$ represents the local objective function at the $i$-th worker.

\subsection{Related Work}

Synchronous \textbf{centralized SGD} with a parameter server is a form of parallel mini-batch SGD. In this approach, workers compute stochastic gradients of their local objectives in parallel and use the averaged gradient to update model parameters after each iteration. The update rule is as follows:

\begin{equation}
x_{k+1} = x_k - \eta \left[ \frac{1}{N} \sum_{i=1}^{N} g_i(x_k;\xi_i) \right],
\end{equation}

where $x_k$ represents the model parameters of server at the $k$-th iteration, $g_i(x_k;\xi_i)$ denotes the gradient at learner $i$ in the $k$-th iteration, $\eta$ is the learning rate, and $\xi_i$ represents randomly sampled mini-batches from the local data distribution. $g_i$ is the gradient estimate at the ith learner in the $k$th iteration calculated using $x_k$. The convergence analysis of this approach has been presented in previous work \cite{dekel2012optimal, bottou2018optimization}. 

\textbf{Centralized SGD} with a parameter server \cite{dean2013tail, li2014communication, cui2014exploiting, dutta2016short, dean2012large} encounters the communication bottleneck problem when the framework has either a large number of workers or low network bandwidth \cite{liu2020decentralized, lian2018asynchronous, lian2017can, wang2021cooperative}. To overcome this bottleneck, \textbf{decentralized SGD} frameworks have been proposed. In the D-PSGD (Decentralized Parallel Stochastic Gradient Descent) approach~\cite{wei2015intelligently}, workers perform one local update and average their models only with neighboring workers. The update rule for D-PSGD is as follows:

\begin{equation}
x_{k+1,i} = \sum_{j=1}^{N} W_{ij}\left[x_{k,j} - \eta g_j(x_{k,j};\xi_{k,j})\right],
\end{equation}

where $x_{k,j}$ denotes the model parameters of worker $j$ at iteration $k$, $\xi_{k,j}$ represents a batch sampled from the local data distribution of worker $j$ at iteration $k$, $W \in \mathbb{R}^{N \times N}$, and $W_{ij}$ is the $(i,j)$-th element of the mixing matrix $W$, which indicates the adjacency between nodes $i$ and $j$. $W_{ij}$ is non-zero if and only if nodes $i$ and $j$ are connected.

The convergence of distributed training is tied to the \textbf{data distribution} across workers. For efficient distributed training, the data partitions at different workers should have similar data distribution. A simple random partitioning in equal-sized partitions may not preserve class-level distribution across partitions. Class-level random partitioning ensures that the number of samples of different classes in any partition is in the same proportion as in the original dataset. However, this may still not guarantee that the feature distribution of a class is similar across different partitions. Ensuring similar feature distributions across different partitions accelerates convergence because it promotes consistent learning, stable gradients, efficient data usage, and better generalization, allowing the model to learn faster and converge more quickly to an optimal solution.

Some recent works involve the use of submodular functions for data partitioning for efficient distributed machine learning. In ~\cite{wei2015intelligently}  using a greedy algorithm involving the use of submodular functions, the dataset was partitioned into subsets with IID features.
However, their greedy algorithm has three major problems: 

\begin{enumerate}
    \item\textbf{Non-uniform sized partitions} The algorithm can lead to different sized partitions. This will lead to a straggler problem in any distributed training algorithms with a synchronization barrier. 

    \item\textbf{Class imbalance across partitions}
    The algorithm can lead to a different number of samples per class across the partitions. This imbalance can lead to lower performance of the aggregated model. The reason why this can happen is that if there are two classes with an overlap in feature space (e.g., Figure~\ref{fig:cat-dog}), then since the greedy algorithm is only maintaining similar feature distribution across partitions, we can have uneven number of samples from these two classes in different partitions.
    \begin{figure}[H]
        \centering
       \includegraphics[scale = 0.5]{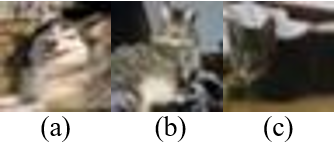}
        \caption{The similarity between dog (a) and cat (b) is higher than similarity between cat (b) and cat (c). Similarity is calculated using cosine similarity and Gaussian kernel with L2 distance. More details about similarity calculation can be found in Section~\ref{sec:format}.}
        \label{fig:cat-dog}
    \end{figure}

    \item\textbf{High computational complexity}: With $n$ samples in the dataset, the greedy algorithm requires $O(n^2)$ computations which can be prohibitive for large datasets. 
\end{enumerate}

The approach was generalized in~\cite{wang2021constrained} to perform constrained submodular partitioning. The experiment demonstrated that the average performance of models trained individually on different subsets is better than random partitioning. However, this was never demonstrated in distributed training setting with communication between workers. In addition, it cannot handle heterogeneous environments, the issues of class imbalance and high complexity remain.


\section{Proposed Method}
\label{sec:format}

In this paper, we propose a novel algorithm called RCD-SGD. RCD-SGD is a meta-scheme algorithm that can be applied to most decentralized SGD algorithms. RCD-SGD includes two parts: data partition and distributed machine learning. Both of them are described in Algorithm 1 and Algorithm 2.

\begin{figure*}[t!]
    \centering
    \includegraphics[scale=0.5]{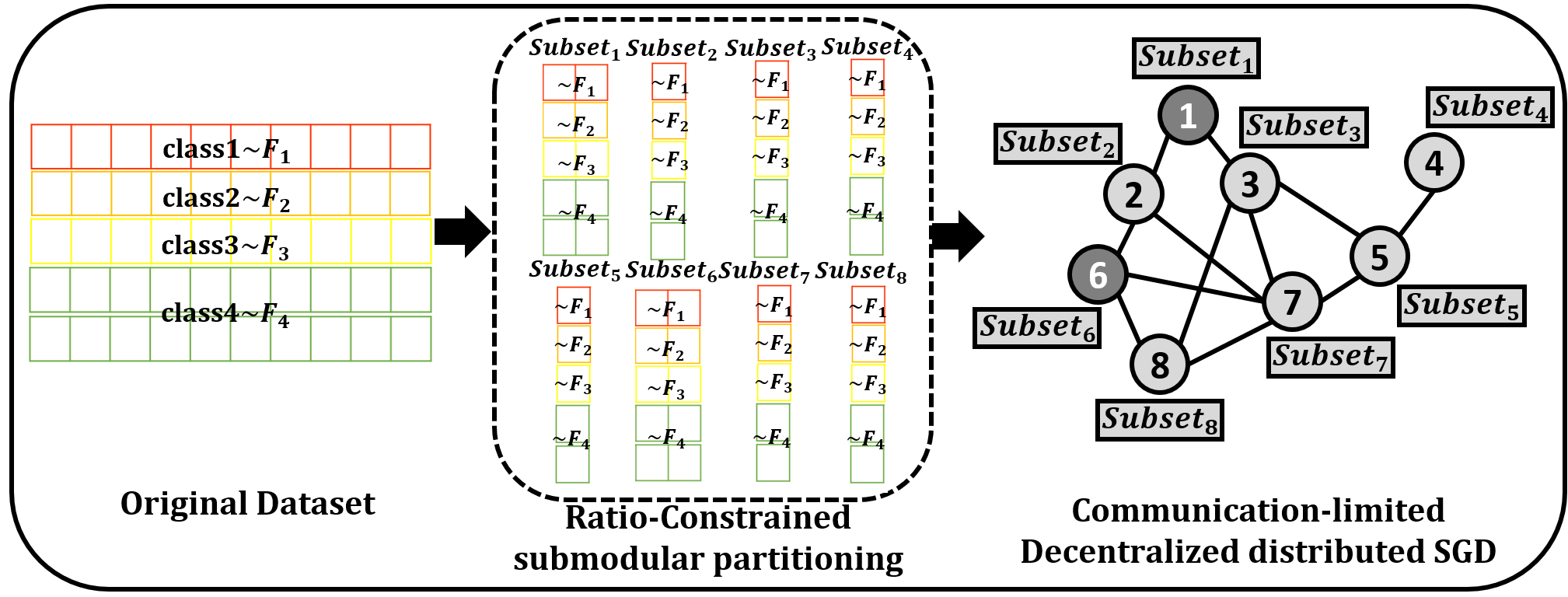}
    \caption{RCD-SGD performs ratio-constrained partitioning of datasets, where the number of samples per class is proportional to the compute capabilities of the workers while maintaining per class feature distribution across partitions. When training multiple epochs of local SGD at workers leads to faster convergence with reduced communication overhead.}
    \label{fig:rcd-sgd}
\end{figure*}

\subsection{Resource-Constrained submodular partitioning}
\begin{algorithm}[t!]
    \begin{algorithmic}[1]
    \caption{Ratio-Constrained submodular partitioning for distributed SGD in heterogeneous environment}
    \label{Alg: 1}
    \STATE \textbf{Initialization:} submodular function $f$, ground set $V$ , number of blocks $N$, number of classes $L$ \;
    \STATE Set the ratio of constraint according to the computational performance of different workers $r_1, r_2,\ldots, r_N$ \;
    \STATE Let $A_1 = A_2 = ... = A_N = \emptyset$
    \STATE Split the ground set $V$ into $L$ sets $V_1, V_2, ..., V_L$ according to class labels.
    \FOR{$V_l = V_1, V_2, ..., V_L$}
        \STATE Constraint $C_{l, j} = \frac{|V_l| r_j}{\sum_{j=1}^N r_j}, j \in \{1,\ldots,N\}$
        \STATE Let $A_1^l = A_2^l = ... = A_N^l = \emptyset, J = [N], R = V_l$ \;
        \WHILE{$R \ne \emptyset$ and $J \ne \emptyset$}
            \STATE $j^* \in$ argmin $_{j\in J} f(A_j^l)$ \;
            \IF{ $\exists v \in R$ s.t. $A_{j^*}^n \cup \{v\} \in C_{l, j^*}$ } 
                \STATE $v^* :=$ GreedyStep($R, C_{l,j^*}, A_{j^*}^l$) \;
                \STATE $A_{j*}^l := A_{j^*}^l \cup \{v^*\}$, $R := R \backslash \{v^*\}$ \;
            \ELSE {} 
                \STATE $J  = J\backslash j^*$ \;
            \ENDIF
        \ENDWHILE
        \STATE Joint subset: $A_1=A_1 \cup A_1^l,..., A_N=A_N \cup A_N^l$
    \ENDFOR
    \STATE \textbf{Output:}($A_1, A_2, A_3,...,A_N$) \;
    \end{algorithmic}
\end{algorithm}

\begin{figure}[t!]
    \centering
   \includegraphics[scale = 0.38]{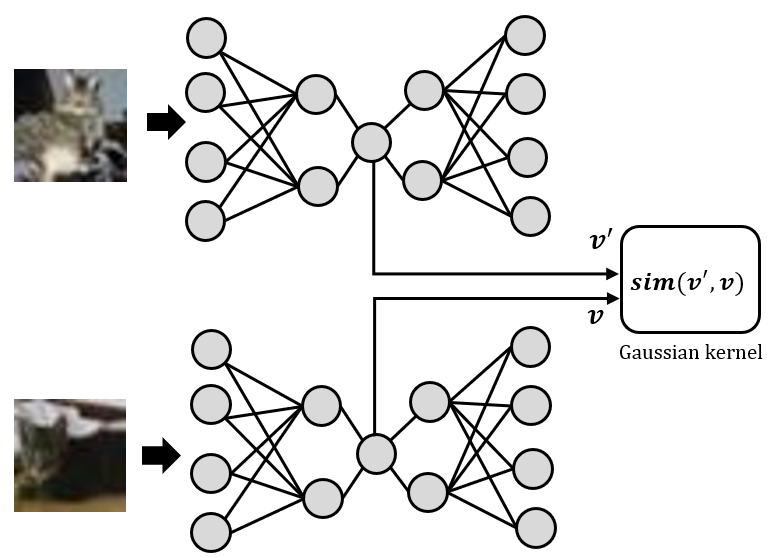}
    \caption{Features of images are extracted from the bottleneck of pre-trained auto-encoder. We use a Gaussian Kernel to measure the similarity.}
    \label{fig:encoder}
\end{figure}

Algorithm~\ref{Alg: 1} is our proposed algorithm for ratio-constrained submodular partitioning. The similarities between each data point are computed using vectors $v$, $v^{'}$. The vectors, which present the features of the encoded image, are extracted as the output of the auto-encoder’s bottleneck. In Figure~\ref{fig:encoder}, we present the process to calculate similarity between two images. More details of the pre-trained auto-encoder model can be found in Section~\ref{sec:exp}. In heterogeneous environments, there can be a wide gap in the compute capabilities of different workers. To solve this problem, we set different constraints in the submodular optimization algorithm according to compute performance of workers to ensure that the number of  samples in a subset is proportional to the performance of the worker training with that subset. (The compute performance of different GPU workers can vary based on factors like GPU model, architecture, and the number of cores, with high-end GPUs offering better performance compared to mid-range ones.) Besides, instead of doing partitioning on the whole dataset, we do class-level partitioning of the dataset. By using consistent constraint across all classes for a certain worker, we can get balanced subclasses. The submodular optimization greedy algorithm requires $O(n^2)$ computations, whereas our algorithm will reduce the computation complexity from $O(n^2)$ to $O(\frac{n^2}{L})$ in $L$ classes dataset thereby achieving $100\times$ speedup when partitioning CIFAR-100.

 \begin{algorithm}[t!]
    \begin{algorithmic}[1]
    \caption{Label-balanced submodular partitioning}
    \label{Alg: 3}
    \STATE \textbf{Initialization:} submodular function $f$, ground set $V$ , number of blocks $N$, number of classes $L$ \;
    \STATE Let $A_1 = A_2 = ... = A_N = \emptyset$
    \STATE Split the ground set $V$ into $L$ sets $V_1, V_2, ..., V_L$ according to class labels.
    \FOR{$V_l = V_1, V_2, ..., V_L$}
        \STATE Constraint $C_l = \frac{|V_l|}{N}$
        \STATE Let $A_1^l = A_2^l = ... = A_N^l = \emptyset, J = [N], R = V_l$ \;
        \WHILE{$R \ne \emptyset$ and $J \ne \emptyset$}
            \STATE $j^* \in$ argmin $_{j\in J} f(A_j^l)$ \;
            \IF{ $\exists v \in R$ s.t. $A_{j^*}^l \cup \{v\} \in C_{l,j^*}$ } 
                \STATE $v^* :=$ GreedyStep($R, C_{l,j^*}, A_{j^*}^l$) \;
                \STATE $A_{j*}^l := A_{j^*}^l \cup \{v^*\}$, $R := R \backslash \{v^*\}$ \;
            \ELSE {} 
                \STATE $J  = J\backslash j^*$ \;
            \ENDIF
        \ENDWHILE
        \STATE Joint subset: $A_1 = A_1 \cup A_1^l,..., A_N = A_N \cup A_N^l$
    \ENDFOR
    \STATE \textbf{Output:}($A_1, A_2, A_3,...,A_N$) \;
    \end{algorithmic}
\end{algorithm}

We formulate the general resources-constrained submodular partitioning optimization task as follows: partition datasets into $N$ groups such that each group contains similar, sufficiently, and approximately-IID strong prediction power. The dataset of class $l$: $\mathbf{V}_l=\{\mathbf{X}, \boldsymbol{y}\}$ is given, where $\mathbf{X}\in \mathbf{R}^{M \times d}$ and $\boldsymbol{y}\in \left\{0,1\right\}^{M}$. Set the constraint according to the computational capability of different workers: $C_{l,1}, C_{l,2}, ..., C_{l,N}$ so that $\sum_{i=1}^N C_{l,i} = M$. Define a partition of $V_l$ as $\{A_1^l, A_2^l,\ldots, A_N^l\}$ where $A_{n}^l\in V_{l}$ for $n=1,2,\ldots , N$ such that 
\begin{equation*}
    A_{j}^l \cap A_{i}^l = \emptyset  \quad   \forall i,j \in \{1,2,\ldots ,N\}, i\ne j 
\end{equation*}
\begin{equation*}
    \bigcup_{n=1}^N A_n^l =  V_{l}
\end{equation*}
The proposed Resource-Constrained submodular partitioning algorithm (Algorithm~\ref{Alg: 1}) addresses the partition task. The $GreedyStep(R, C_l, A_{j^*}^l)$ in Algorithm~\ref{Alg: 1}  utilizes the submodular function to measure the value. The GreedyStep is defined as:

\begin{equation*}
    v^*= \arg\max_{i\in {V_l\backslash \hat{V_{l}}}}f\left({A_{j}^l}\cup \left\{v^i\right\}\right)
\end{equation*}
where $\bigcup_{n=1}^N A_n^l = \hat{V_l} \subset V_l$. After partition the subsets will finally achieve: $\bigcup_{n=1}^N A_n^l = V_l$. We define a discrete set function $f\colon \mathbf{V}\rightarrow \mathbf{R}$ as a submodular function if
\begin{equation*}
   \resizebox{0.8\hsize}{!}{$f\left({\hat{A_j^l}\cup \left\{v^*\right\}}\right) - f\left(\hat{A_j^l}\right) \geq f\left({A_j^l\cup \left\{v^*\right\}}\right) - f\left(A_j^l\right)$}
\end{equation*}
where $\hat{A_j^l} \subset A_j^l, \forall j\in \{V_l \backslash \cup_{n=1}^N A_j^l\}$. A submodular function $f$ must be monotone non-decreasing, which is the basic requirment:
\begin{equation*}
    f\left({A_j^l\cup \left\{v^*\right\}}\right) - f\left(A_j^l\right) \geq 0
\end{equation*}

When the performance of the workers are similar, the Algorithm~\ref{Alg: 1} can also be simplified to label-balanced submodular partitioning, which is presented in Algorithm~\ref{Alg: 3}.

\subsection{Distributed SGD}


We next propose a synchronous distributed training algorithm using our proposed partitioning algorithm. The main steps are depicted in Figure~\ref{fig:rcd-sgd}. Partitioning the data as in Algorithm~\ref{Alg: 1} ensures that each worker gets total (and per class) samples proportional to their processing speed. This will achieve approximate IID partitioning across workers, both at the feature level and label level. After partitioning the data, we perform synchronous distributed training with several rounds of local SGD before each synchronization step. This will not be detrimental to our convergence as we are ensuring IID partitions.

Our training algorithm is a modification of Distributed-Parallel SGD (D-PSGD)~\cite{wei2015intelligently} where the data partitions can be Non-IID due to random partitioning. The Non-IIDness can be in feature and/or label space. Non-IIDness in feature refers that features or labels of the data used in a machine learning model are not independently and identically distributed. Thus convergence can be poor with D-PSGD, especially for datasets with an overlap in feature space across classes. Further D-PSGD has a communication barrier after every step of local training and hence takes a longer time to converge. 
\begin{algorithm}[t!]
    \begin{algorithmic}[1]
    \caption{Decentralized distributed SGD algorithm}
    \label{Alg: 2}
    \STATE \textbf{Initialization:} initialize local models $\{x_0^i\}_{i=1}^N$ with the same initialization, the ratio of computational performance for different workers is $r_1, r_2, ..., r_N$, learning rate $\gamma$, batch size $ B\frac{r_1}{r_i}$, communication frequency $F$, weight matrix $W$, and the total number of iterations $K$. Import the local subset $\xi_{i}$ from subsets $\{ A_{1}, A_{2}, ..., A_{N} \}$, which is partitioned by Resource-Constrained submodular partitioning algorithm\;
    \WHILE{$k=0, 1, 2, ... K-1$}
        \STATE Compute  the local stochastic gradient $\nabla F_i(x_{k, i}; \xi_{k, i})$ on all nodes\;
        \IF{$k \% F$  == 0}
            \STATE Compute the neighborhood weighted average by fetching neighbor models: $\hat{x}_{k, i} = \sum_{j=1}^N W_{ij}x_{k,j}^b$\;
        \ELSE {} 
            \STATE $\hat{x}_{k, i} = x_{k, i}$ \;
        \ENDIF
        \STATE Update local model: $x_{k+1, i} = \hat{x}_{k, i} - \gamma \nabla F_i(x_{k, i}; \xi_{k, i})$\;
    \ENDWHILE
    \STATE \textbf{Output:} Average of all workers $\frac{1}{N} \sum_{i=1}^N x_{K-1,i}$\;
    \end{algorithmic}
\end{algorithm}
During training, the communication between workers happens over a computational graph where the nodes represent the workers and the edges denote the connection between the workers. Thus $i$th, $j$th component of $W$, $W_{ij}$, is non-zero if and only if node $i$ and node $j$ are connected. The workers only communicate after $F$ iterations of local SGD. Since the batch size and the dataset shard per worker is proportional to its processing speed, our partitioning ensures that all the workers take approximately same amount of time to finish $F$ iterations locally thereby reducing the straggler problem with synchronous SGD.

\begin{figure*}[t!]
    \centering
    \includegraphics[width=180mm]{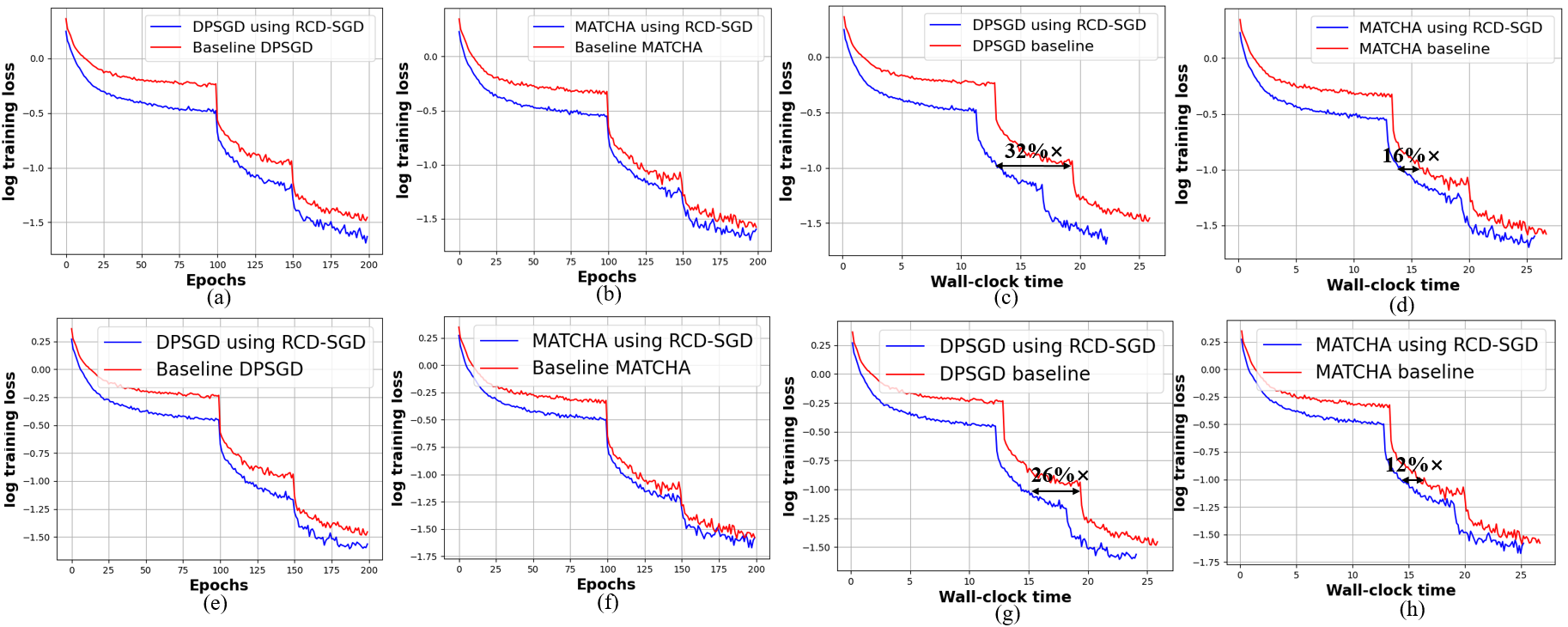}
    \caption{RCD-SGD use facility location in $greedystep$: (a, b, c, d); use graph cut $greedystep$: (e, f, g, h). Results were obtained on CIFAR-10 data set using ResNet-50. (a), (b) and (e), (f) show convergence with number of epochs while (c), (d) and (g), (h) show convergence with wall clock time.}
    \label{Fig: 3}
\end{figure*}

\begin{figure*}[t!]
    \centering
    \includegraphics[width=180mm]{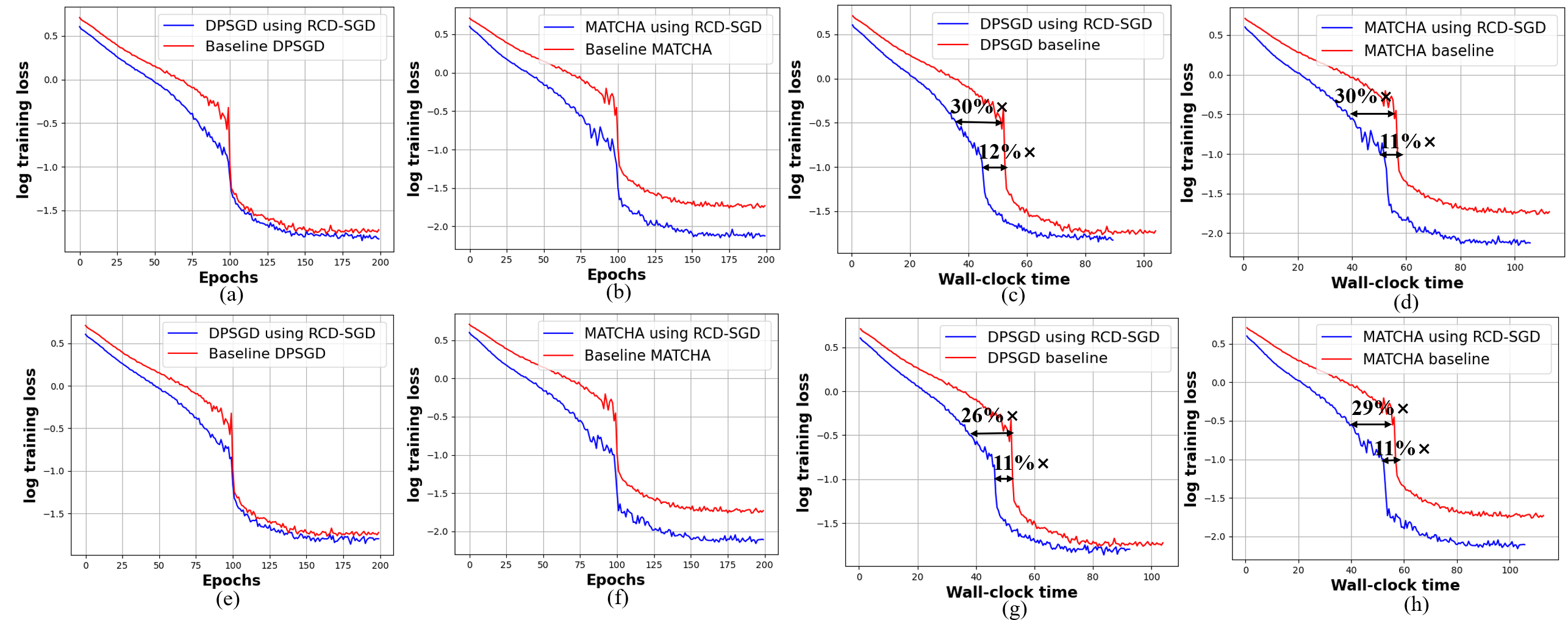}
    \caption{RCD-SGD use facility location in $greedystep$: (a, b, c, d); use graph cut $greedystep$: (e, f, g, h). Results were obtained on CIFAR-100 data set using WideResNet. (a), (b) and (e), (f) show convergence with number of epochs while (c), (d) and (g), (h) show convergence with wall clock time.}
    \label{Fig: 4}
\end{figure*}

\section{Experiments}
\label{sec:exp}

In Fig.~\ref{Fig: 3} and Fig.~\ref{Fig: 4}, we report the results of RCD-SGD using D-PSGD and MATCHA as baselines. Performance of RCD-SGD using facility location and graph cut as submodular functions are included. Results are compared with baseline D-PSGD and MATCHA. MATCHA, like D-PSGD, begins with a predetermined communication network topology. However, unlike D-PSGD, MATCHA allows the system designer to define a flexible communication budget $c_b$, representing the average communication frequency over network links. The dynamic construction of the weight matrices ${W^{(k)}}$ is dependent on the chosen budget $c_b$. When $c_b$ = 1, MATCHA is equivalent to the vanilla D-PSGD algorithm. If $c_b$ is less than 1, MATCHA effectively decreases the communication frequency over each link, taking into consideration the link's significance in maintaining the overall graph connectivity. Furthermore, MATCHA assigns probabilities to worker connections, enabling their activation in certain iterations. Experiments use the following setting:\\
\begin{enumerate}
    \item\textbf{Models, dataset, and compared algorithms:} The performance of all algorithms is evaluated on image classification task using CIFAR-10 and CIFAR-100 ($|V|$ = 500000) ~\cite{krizhevsky2009learning}. In our experiments, we employ ResNet-50~\cite{he2016deep} and Wide ResNet models~\cite{zagoruyko2016wide}. We implement RCD-SGD as modification of D-PSGD and MATCHA with a communication budget $c_b = 0.5$. \\
    \item\textbf{Submodular functions:} We use facility location, i.e., $f(A_n^l) = \sum_{v\in V_l} max_{v^{\prime} \in A_n^l} sim(v, v^{\prime})$ and Graph Cut, i.e., $f(A_n^l) = \sum_{v\in V_l \backslash A_n^l} \sum_{v^{\prime} \in A_n^l} sim(v, v^{\prime})$ in $Greedystep$ (algorithm 1) separately to run the experiments. $sim(v, v^{\prime})$ is the similarity between $(v, v^{\prime})$. We use a Gaussian kernel with L2 distance to measure the similarity. We set Gaussian Kernel as $\sigma = \sum_{v, v^{'} \in V} {{||v - v^{'}||}_{2}} / n^2$, where $\sigma$ is the bandwidth of the kernel. Similarities between each data point are computed by vectors $v, v^{'}$. In CIFAR-10 and CIFAR-100, the vectors are obtained from the bottleneck layer’s outputs of a deep auto-encoder model. The partition step is not included in the comparison of wall-clock times, since the partition only needs to be done once to generate subsets.\\
    \item\textbf{Implementations:} All algorithms are trained for a sufficiently long time until convergence or onset of over-fitting. The learning rate is fine-tuned for the D-PSGD baseline and then used for all other algorithms. We set the initial learning rate as 0.8 and it decays by 10 after 100 and 150 epochs. The batch size per worker node is 64. RCD-SGD uses $F = 2$ and reduces the communication frequency to 50\%. The auto-encoder is trained using ReLU non-linearity and batch normalization. The network is trained in PyTorch using the procedure described in \cite{lavania2019auto, wang2021constrained}. The auto-encoder is pre-trained using image reconstruction task. The neural network architecture of auto-encoder can be found in table~\ref{tab:1}. The auto-encoder utilizes ADAM as the optimization method. The initial learning rate of 5e-3, with a weight decay of 5e-4 and a minibatch size of 100.

\begin{table}[h]
    \centering
    \resizebox{\linewidth}{!}{
    \begin{tabular}{lllllllll}
    \multicolumn{1}{c|}{\textbf{Group}} & \multicolumn{2}{c|}{\textbf{\makecell[c]{Block Type\\(kernel size, stride, channels)}}} &\multicolumn{3}{c}{\textbf{Blocks}}\\ \hline
    \multicolumn{1}{c|}{conv1}     & \multicolumn{2}{c|}{$[3 \times 3], 2, 64$}     &\multicolumn{3}{c}{1}   \\ 
    \multicolumn{1}{c|}{conv1 (residual)}     & \multicolumn{2}{c|}{$\begin{bmatrix} 3 \times 3\\3 \times 3\\\end{bmatrix}, 1, 64$}     &\multicolumn{3}{c}{2}   \\ 
    \multicolumn{1}{c|}{conv2}     & \multicolumn{2}{c|}{$[3 \times 3], 2, 16$}     &\multicolumn{3}{c}{1}   \\ 
    \multicolumn{1}{c|}{conv2 (residual)}     & \multicolumn{2}{c|}{$\begin{bmatrix} 3 \times 3\\3 \times 3\\\end{bmatrix}, 1, 16$}     &\multicolumn{3}{c}{2}   \\ 
    \multicolumn{1}{c|}{conv3}     & \multicolumn{2}{c|}{$[3 \times 3], 2, 8$}     &\multicolumn{3}{c}{1}   \\ 
    \multicolumn{1}{c|}{conv3 (residual)}     & \multicolumn{2}{c|}{$\begin{bmatrix} 3 \times 3\\3 \times 3\\\end{bmatrix}, 1, 8$}     &\multicolumn{3}{c}{2}   \\ 
    \multicolumn{1}{c|}{conv4}     & \multicolumn{2}{c|}{$[3 \times 3], 1, 4$}     &\multicolumn{3}{c}{1}   \\ 
    \multicolumn{1}{c|}{conv4 (residual)}     & \multicolumn{2}{c|}{$\begin{bmatrix} 3 \times 3\\3 \times 3\\\end{bmatrix}, 1, 4$}     &\multicolumn{3}{c}{1}   \\
    \multicolumn{1}{c|}{deconv4 (residual)}     & \multicolumn{2}{c|}{$[3 \times 3], 1, 4$}     &\multicolumn{3}{c}{1}   \\ 
    \multicolumn{1}{c|}{deconv3}     & \multicolumn{2}{c|}{$\begin{bmatrix} 3 \times 3\\3 \times 3\\\end{bmatrix}, 1, 8$}     &\multicolumn{3}{c}{1}   \\
    \multicolumn{1}{c|}{deconv3 (residual)}     & \multicolumn{2}{c|}{$\begin{bmatrix} 3 \times 3\\3 \times 3\\\end{bmatrix}, 1, 9$}     &\multicolumn{3}{c}{2}   \\ 
    \multicolumn{1}{c|}{deconv2}     & \multicolumn{2}{c|}{$\begin{bmatrix} 3 \times 3\\3 \times 3\\\end{bmatrix}, 2, 16$}     &\multicolumn{3}{c}{1}   \\
    \multicolumn{1}{c|}{deconv2 (residual)}     & \multicolumn{2}{c|}{$[3 \times 3], 2, 16$}     &\multicolumn{3}{c}{2}   \\ 
    \multicolumn{1}{c|}{deconv1}     & \multicolumn{2}{c|}{$[3 \times 3], 2, 64$}     &\multicolumn{3}{c}{1}   \\
    \multicolumn{1}{c|}{deconv1 (residual)}     & \multicolumn{2}{c|}{$\begin{bmatrix} 3 \times 3\\3 \times 3\\\end{bmatrix}, 2, 64$}     &\multicolumn{3}{c}{2}   \\ 
    \multicolumn{1}{c|}{deconv0}     & \multicolumn{2}{c|}{$[3 \times 3], 2, 3$}     &\multicolumn{3}{c}{1}   \\ \hline \\
    \end{tabular}}
    \caption{\label{tab:widgets} Neural network architecture of the Auto-encoder.}
    \label{tab:1}
\end{table}

    \item\textbf{Machines and clusters: }All the implementations are compiled with PyTorch and OpenMPI within mpi4py and rtx8000 GPUs as workers. We conduct experiments on a HPC cluster with 100Gbit/s Infini-band network. \\
\end{enumerate}

We conducted an analysis of the results obtained with RCD-SGD. In both Figure~\ref{Fig: 3} and Figure~\ref{Fig: 4}, we observed that RCD-SGD, utilizing either facility location or graph cut as submodular functions, consistently outperformed the baseline methods. Notably, improvements were evident in both the convergence rate over epochs and the convergence speed with respect to wall-clock time.

When training the results from CIFAR-10 and ResNet-50, RCD-SGD achieved a significant time saving of up to 30\% (Fig.\ref{Fig: 3} c). Similarly, when utilizing CIFAR-100 and WideResNet, substantial performance improvements were observed, with a time saving of up to 12\% (Fig.\ref{Fig: 4} c), measured until the log of training loss reached -1.0.

By analyzing the results, it becomes evident that RCD-SGD not only accelerates the convergence speed of both baselines but also demonstrates a remarkable improvement in convergence. The use of IID subsets aids in achieving faster convergence for RCD-SGD, as the localized training process enables more efficient model updates and fosters enhanced learning across the dataset. This advantage can be attributed to the network's ability to focus on specific patterns and features within each subset, leading to better training dynamics and overall performance.

An interesting aspect of RCD-SGD is that it maintains convergence even after the baselines have already converged, thanks to the independent and identically distributed (IID) partitioned subsets. At the 100th epoch, before the learning rate decay, RCD-SGD exhibited up to 30\% wall-clock time saving (Fig.~\ref{Fig: 4} c) when measured until the log of training loss reached -0.05.

Furthermore, the final loss and test accuracy were consistently improved when employing RCD-SGD, as demonstrated in Table~\ref{tab:2}. It is important to note that all the results presented in Table~\ref{tab:2} are the average of ten experiments, ensuring robustness and reliability.


\begin{table}[h]
\centering
\resizebox{\linewidth}{!}{
\begin{tabular}{lllllllll}
\multicolumn{1}{c|}{\textbf{Dataset}}                           &\multicolumn{1}{c|}{\textbf{Model}}       &\multicolumn{1}{c|}{\textbf{Algorithms}}       & \multicolumn{1}{c}{\textbf{Accuracy}}\\ \hline
\multicolumn{1}{c|}{\multirow{4}{*}{\textbf{CIFAR-100}}}        &\multicolumn{1}{c|}{\multirow{4}{*}{\textbf{WideResNet}}}       &\multicolumn{1}{c|}{D-PSGD}                    & \multicolumn{1}{c}{0.718} \\ 
\multicolumn{1}{c|}{~}                                          &\multicolumn{1}{c|}{~}       &\multicolumn{1}{c|}{D-PSGD based RCD-SGD}      & \multicolumn{1}{c}{\textbf{0.752}} \\
\multicolumn{1}{c|}{~}                                          &\multicolumn{1}{c|}{~}       &\multicolumn{1}{c|}{MATCHA}                    & \multicolumn{1}{c}{0.755} \\
\multicolumn{1}{c|}{~}                                          &\multicolumn{1}{c|}{~}       &\multicolumn{1}{c|}{MATCHA based RCD-SGD}      & \multicolumn{1}{c}{\textbf{0.762}}   \\ \hline
\multicolumn{1}{c|}{\multirow{4}{*}{\textbf{CIFAR-10}}}         &\multicolumn{1}{c|}{\multirow{4}{*}{\textbf{ResNet-50}}}       &\multicolumn{1}{c|}{D-PSGD}                    & \multicolumn{1}{c}{0.925}  \\
\multicolumn{1}{c|}{~}                                          &\multicolumn{1}{c|}{~}       &\multicolumn{1}{c|}{D-PSGD based RCD-SGD}      & \multicolumn{1}{c}{\textbf{0.937}}   \\
\multicolumn{1}{c|}{~}                                          &\multicolumn{1}{c|}{~}       &\multicolumn{1}{c|}{MATCHA}                    & \multicolumn{1}{c}{0.931} \\
\multicolumn{1}{c|}{~}                                          &\multicolumn{1}{c|}{~}       &\multicolumn{1}{c|}{MATCHA based RCD-SGD}      & \multicolumn{1}{c}{\textbf{0.939}}   \\ \hline   \\
\end{tabular}}
\caption{\label{tab:widgets} Test accuracy obtained with MATCHA and MATCHA-based AL-DSGD for ResNet-50 model trained on CIFAR-10 and WideResNet model trained on CIFAR-100.}
\label{tab:2}
\end{table}

\section{Conclusion}
\label{sec:typestyle}
Distributed training in heterogeneous clusters requires efficient data partitioning for faster convergence. RCD-SGD achieves IID partitioning with similar per-class feature distribution across workers having different compute capabilities. The training can be performed with increased epochs of local training leading to reduced synchronization overhead. We are exploring use of other submodular functions and the sensitivity of distributed SGD convergence on their choice. 


{\small
\bibliographystyle{ieee_fullname}
\bibliography{egpaper_final}
}

\end{document}